\documentclass[10pt,twocolumn,letterpaper]{article}

\usepackage[accsupp]{axessibility}  

\usepackage{cvpr}             

\definecolor{cvprblue}{rgb}{0.21,0.49,0.74}
\usepackage[pagebackref,breaklinks,colorlinks,allcolors=cvprblue]{hyperref}
\usepackage{colortbl}

\def\paperID{4} 
\def\confName{CVPR}
\def\confYear{2025}

\newcommand{\cmark}{\ding{51}}%
\newcommand{\xmark}{\ding{55}}%

\title{Window Token Concatenation for Efficient Visual Large Language Models}

\author{Yifan Li\textsuperscript{1}, Wentao Bao\textsuperscript{1}, Botao Ye\textsuperscript{2}, Zhen Tan\textsuperscript{3}, Tianlong Chen\textsuperscript{4}, Huan Liu\textsuperscript{3}, Yu Kong\textsuperscript{1}\\
{\normalsize \textsuperscript{1} {Michigan State Univerisity, \texttt{\{liyifa11, baowenta, yukong\}@msu.edu}}}\\
{\normalsize 428 SouthShaw Lane, East Lansing, MI 48824, USA}\\
{\normalsize \textsuperscript{2} ETH Zürich, \texttt{botao.ye@inf.ethz.ch}}\\
{\normalsize ETH Zürich, Rämistrasse 101, Zürich 8092, Switzerland} \\
{\normalsize \textsuperscript{3} Arizona State University, \texttt{\{ztan36, huanliu\}@asu.edu}}\\
{\normalsize Arizona State University, 1151 S Forest Ave, Tempe, AZ 85281, USA}\\
{\normalsize \textsuperscript{4} University of North Carolina at Chapel Hill, \texttt{tianlong@cs.unc.edu}} \\
{\normalsize UNC Chapel, 145E. Cameron Street, Hill Hall, Chapel, NC 27514, USA}
}

\begin{document}
\maketitle
\begin{abstract}
  To effectively reduce the visual tokens in Visual Large Language Models (VLLMs), we propose a novel approach called \textbf{Wi}ndow Token \textbf{Co}ncatenation (\textbf{WiCo}). Specifically, we employ a sliding window to concatenate spatially adjacent visual tokens. However, directly concatenating these tokens may group diverse tokens into one, and thus obscure some fine details. To address this challenge, we propose fine-tuning the last few layers of the vision encoder to adaptively adjust the visual tokens, encouraging that those within the same window exhibit similar features.
  To further enhance the performance on fine-grained visual understanding tasks, we introduce WiCo$+$, which decomposes the visual tokens in later layers of the LLM.
  Such a design enjoys the merits of the large perception field of the LLM for fine-grained visual understanding while keeping a small number of visual tokens for efficient inference.  We perform extensive experiments on both coarse- and fine-grained visual understanding tasks based on LLaVA-1.5 and Shikra, showing better performance compared with existing token reduction projectors. The code is available: \url{https://github.com/JackYFL/WiCo}.
\end{abstract}

\section{Introduction}

Large Language Models (LLMs) \cite{radford2019language,ouyang2022training,touvron2023llama}, featuring billions of parameters and trained on vast corpora using an auto-regressive strategy, exhibit impressive performance across a variety of tasks~\cite{tan2024large,zhao2025scale}.  To enhance the capabilities of LLMs operating across multiple modalities \cite{ghosh2024exploring}, researchers are increasingly focusing on Multi-modal Large Language Models (MLLMs), particularly on Visual Large Language Models (VLLMs \cite{li2025visual}) which append a series of visual tokens ahead of {textual} instruction ones \cite{liu2024visual,liu2023improved,zhu2024minigpt}. 


Due to limited computation resources in real world and redundancy inherently in visual images \cite{bolya2023token}, it is desired to reduce the visual tokens for VLLMs' training and inference~\cite{li2024facial}, especially for high-resolution images \cite{liu2024llavanext,li2024mini,dong2024internlm,luobeyond}, videos \cite{zhang2023video,li2023videochat} and multi-image tasks \cite{jiang2024mantis}. 
Visual tokens comprise a major fraction of the total input tokens for an LLM. For instance, a 336$\times$336 resolution image encoded by CLIP ViT-L/336 \cite{radford2021learning} leads to 576 prefix visual tokens, compared to merely around 40$\sim$50 textual instruction tokens. As a result, the computation cost associated with visual tokens is substantial. The way to \textit{effectively reducing visual tokens without adversely affecting the performance 
}presents a significant challenge for VLLMs.
\begin{figure*}[t]
    \centering
    \subcaptionbox{Visual token projectors. \label{fig:proj}}{
        \includegraphics[height=1.5in]{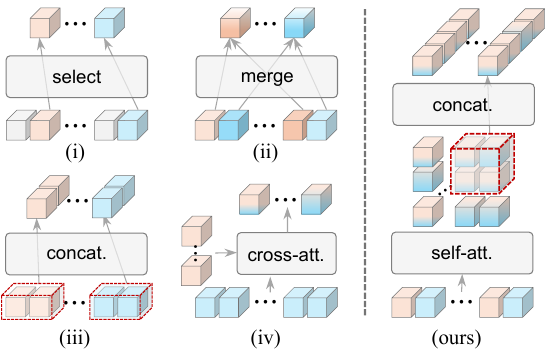}
    }
    \hspace{20pt}
    \subcaptionbox{Task granularity.\label{fig:grain}}{
        \includegraphics[height=1.4in]{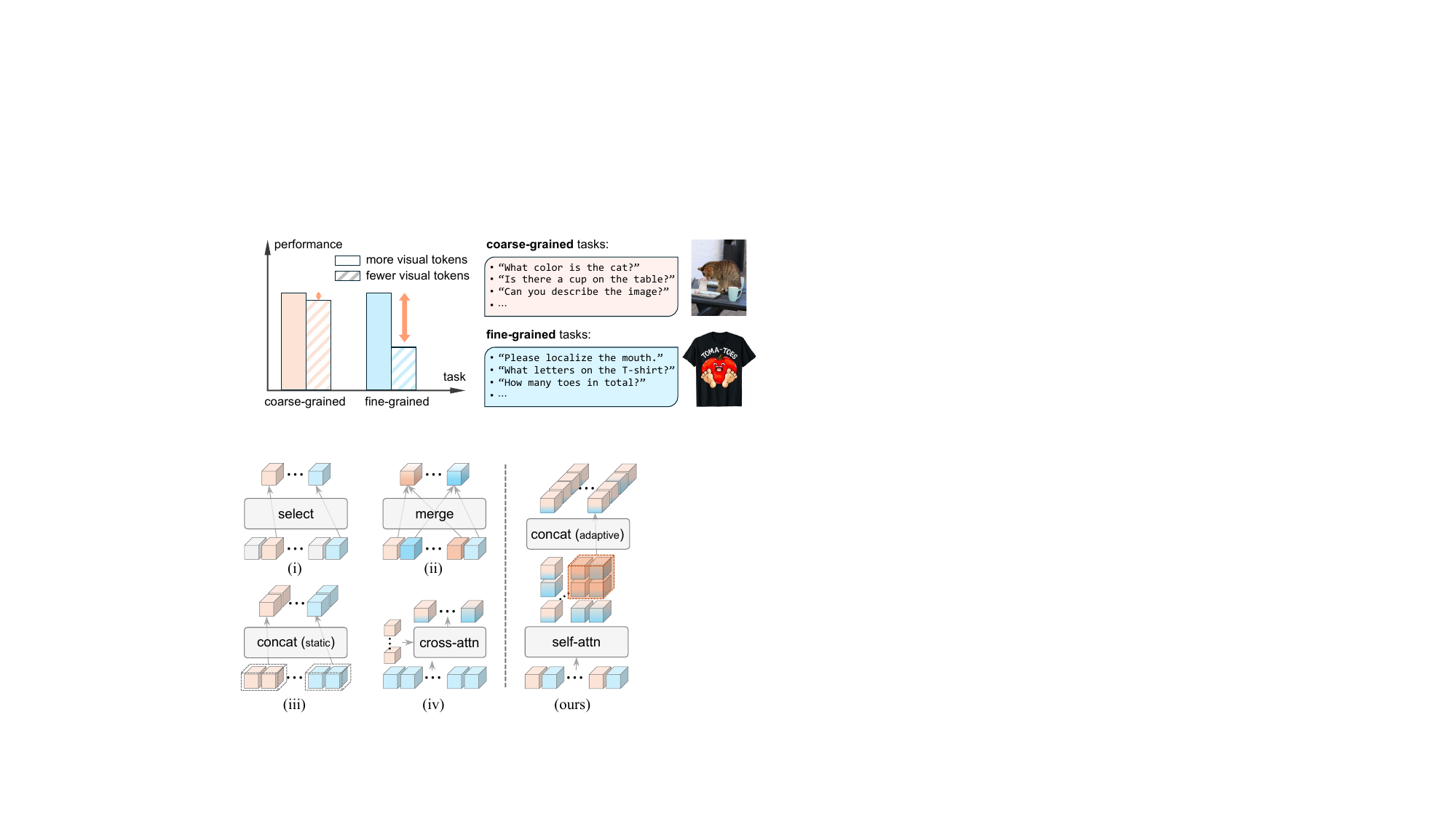}
    }
    \captionsetup{font=normalsize,aboveskip=3pt}
    \caption{Motivations of our method. (a) Current projector types (left) and ours (right) for VLLM token reduction. Existing token reduction projectors are mainly based on (i) selection, (ii) merging, (iii) concatenation and (iv) cross-attention. (b) illustrates that the performance of VLLMs is sensitive to the types of downstream tasks when changing the number of visual tokens. Specifically, the performance of VLLMs will decrease more for fine-grained understanding tasks compared to the course-grained ones when reducing the visual tokens.}
\end{figure*}

Prior work has proposed various visual projectors to reduce the prefix visual tokens, which can be summarized into the four categories as shown in \cref{fig:proj}, \ie, cross-attention resampler \cite{alayrac2022flamingo,li2023blip}, token selection~\cite{liu2024textmonkey}, token merging \cite{cha2023honeybee, tolstikhin2021mlp}, and token concatenation \cite{chen2024minigpt}.
{Despite their promising} performance in various downstream tasks, they still {face} limitations like inflexibility of controlling the output tokens
\cite{chen2024minigpt} and information loss \cite{alayrac2022flamingo,li2023blip,liu2024textmonkey,cha2023honeybee}. Furthermore, these methods do not consider the influence of visual tokens on different types of visual tasks (see \cref{fig:grain}). Specifically, they overlook the distinction between coarse-grained semantic understanding tasks, such as the polling questions in visual question answering (VQA)  \cite{li2023evaluating,li2023blip,liu2024visual}, 
and fine-grained understanding tasks such as object detection/segmentation and optical character recognition (OCR) \cite{chen2023shikra,lai2023lisa,you2024ferret}. This oversight further restricts the generalization capabilities of these methods. Instead, we empirically found that the tasks with different levels of granularity have different sensitivities to the number of visual tokens (\eg, \cref{tab:general}, \cref{tab:grounding}, \cref{fig:k_vqat}). 

To address the aforementioned limitations, 
we introduce a novel approach named Window patch Concatenation (WiCo), and its enhanced version WiCo$+$.
Specifically, we dynamically adjust the visual tokens by tuning the last few layers of the vision encoder and then concatenate the tokens within a 2D sliding window. The benefits are twofold. First, compared with the prevailing methods~\cite{tolstikhin2021mlp,alayrac2022flamingo,li2023blip} that average over the tokens, our token concatenation keeps minimal information loss of the raw visual tokens for LLM decoding. Second, different from  MiniGPTv2~\cite{chen2024minigpt} that defines 1D window for concatenation, our 2D window is intuitively advantageous because that the visual tokens are spatially correlated rather than limited to a single direction~\cite{tian2024visual}, so it can better exploit the spatial locality information. 
Similar to the clustering problem, representations within a window are expected to be similar, while representations across different windows should be distinct.
However, spatial adjacent visual tokens in a window may include significantly different visual features, concatenating them into one may obscure some fine details. To make the features within a window more similar and those across different windows more distinct, we propose dynamically adjusting the visual tokens by fine-tuning the last few layers of the vision encoder (see \cref{fig:tuned_frozen}).

Furthermore, our experiments reveal that \textit{fine-grained understanding tasks are more sensitive to the number of visual tokens compared to the course-grained semantic understanding tasks in VLLMs}. To tackle this problem, {in WiCo$+$}, we propose to decompose the visual tokens in the later decoder layers of the LLM to facilitate the fine-grained understanding tasks. This method intrinsically performs hierarchical attention modeling over the visual patches in token feature space, \ie, window-level attention in early LLM layers and patch-level attention in late LLM layers. 
Our contributions are three-fold:

\setlist[itemize]{leftmargin=*}
\begin{itemize}
    \item We systematically explore the design choices of efficient visual projectors in VLLMs, which significantly impact the performance of the fine-grained visual understanding tasks when reducing the visual tokens;
    \item We introduce a novel visual projector WiCo by dynamic 2D window token concatenation, which enables efficient instruction tuning of VLLMs. Moreover, an enhanced version WiCo$+$ by upsampling visual tokens in later layers of the LLM decoder is proposed to further facilitate the fine-grained visual understanding tasks;
    \item We conduct extensive experiments on various downstream tasks, including general VQA tasks and fine-grained grounding tasks based on LLaVA-1.5 and Shikra. Multiple visual token reduction projector baselines are reproduced to compare with our WiCo ($+$). The results demonstrate the superiority and effectiveness of our method {in terms of both efficiency and efficacy}.
\end{itemize}

\section{Related work}

\subsection{Token reduction in VLLM projector}
\begin{figure*}[t]
    \centering
    \includegraphics[width=0.9\linewidth]{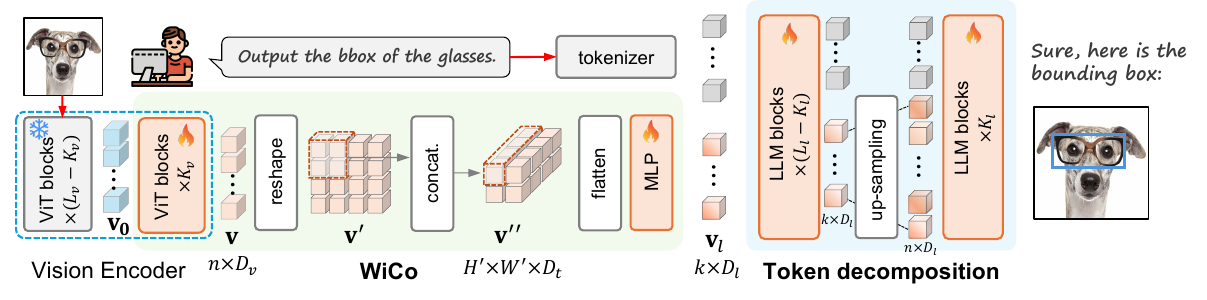}
    \caption{Framework of our WiCo$+$. WiCo$+$ consists of two main components, \ie, a dynamic window token concatenation projector (WiCo) and the token decomposition strategy in the later layers of the LLM decoder. WiCo first learns similar local token representations by $k_v$ self-attention layers from the last $k_v$  layers of a pretrained vision encoder (say CLIP). Then, a sliding window is adopted on the 2-D token map to perform concatenation, and an MLP is utilized to project these visual tokens into language space. To further enhance the perception field of the rest visual tokens, we decompose the visual tokens in the later layers (say the last $K_l$ layers) of the LLM decoder, which will benefit the fine-grained understanding tasks.}
    \label{fig:framework} 
\end{figure*}

VLLMs have shown superior capability on sophisticated visual reasoning tasks. However, the increased number of visual tokens in high-definition images or videos inevitably leads to significant computational costs in the LLM decoder~\cite{dao2022flashattention}. To address this issue, several approaches have been proposed to reduce the number of visual tokens from different perspectives~\cite{alayrac2022flamingo, liu2024textmonkey}.
VLLMs have shown superior capability on sophisticated visual reasoning tasks, \eg, reasoning over text-rich document images~\cite{singh2019towards} and grounding text concepts to pixel locations on images~\cite{kazemzadeh2014referitgame,yu2016modeling}. These tasks typically deal with fine-grained visual details in high-definition images or videos. Unfortunately, the increased visual tokens in high-definition images inevitably result in a huge computational cost in the LLM decoder~\cite{dao2022flashattention}. 

To reduce the number of visual tokens, early works such as the Perciever~\cite{alayrac2022flamingo}/Q-former projector~\cite{li2023blip} use a small number of learnable query tokens to summarize visual content. In~\cite{liu2024textmonkey}, the visual token similarity is utilized to filter out redundant tokens, and cross attention is used to compensate for the information loss of the filtering. In MiniGPT-v2~\cite{chen2024minigpt}, visual tokens are reduced by a simple concatenation over adjacent tokens. To enable flexibility of the number of tokens and the locality of dense visual tokens, recent work~\cite{cha2023honeybee} explores different ``Abstractors'' as visual projectors for the effective token reduction, that uses adaptive average pooling in ResNet blocks (C-Abstractor) and deformable attention blocks (D-Abstractor). 
More recent work~\cite{chen2024image, shang2024prumerge,zhang2024sparsevlm} focuses on improving inference efficiency without introducing extra training components. Fast-V \cite{chen2024image} finds the sparsity inherent in the attention scores from deeper language layers, and proposes to prune the unimportant visual tokens after the certain layer of the LLM. LLaVA-Prumerge \cite{shang2024prumerge} adopts the prior knowledge in vision encoder, \textit{i.e.}, the attention scores calculated by the CLS token and other visual tokens, to prune redundant ones. To preserve the information of the rest tokens,  it merges less informative tokens into the cluster center tokens. SparseVLM \cite{zhang2024sparsevlm} proposes a language-guided selection by utilizing the semantic information in prompts to select related tokens. Compared to these prior arts, our method takes advantage of the flexibility and locality from~\cite{cha2023honeybee} by dynamic window design, and our 2D window concatenation that considers the bidirectional local proximity of visual tokens, which is more effective than the unidirectional concatenation in~\cite{chen2024minigpt}. In our paper, we mainly focus on  designing a visual token reduction projector that preserves as much information as possible. As a result, we do not compare with these baselines in our experiments.

\subsection{Token reduction in vision transformer}
The Vision Transformer (ViT) \cite{attention} is extensively utilized in numerous vision tasks \cite{deit3, mae, dpt, vitdet, ostrack}, but it has long struggled with quadratic complexity. To mitigate computational costs, various token reduction methods have been proposed~\cite{dynamicvit, avit, evit, spvit, ostrack, adavit, bolya2023token}.
Earlier works progressively identify and discard uninformative tokens layer-by-layer~\cite{dynamicvit, avit}, which may lead to information loss. Consequently, more recent approaches either fuse unimportant tokens together~\cite{evit, spvit} or combine semantically similar tokens~\cite{bolya2023token}. However, these methods are primarily designed for situations with pure image input, gradually reducing the number of tokens as the ViT deepens. 
These approaches are less effective for VLLMs, where the ViT serves as an image feature extractor, and the resulting features are fed into vision-language interaction modules. Consequently, reducing features too early causes significant information loss for vision-language interaction. In this paper, we mainly consider reducing the image tokens in VLLM projectors after obtaining all of the visual tokens from the vision encoder.


\section{Method}

The overall framework of our method WiCo+ is illustrated as \cref{fig:framework}.  We will first formulate the problem and then delineate each component in the following subsections.

\subsection{Problem formulation}
For a given image, a vision encoder is utilized to project it into $n$ visual tokens $\mathbf{v}=\{v_1, v_2, ..., v_n\}\in \mathbb{R}^{n\times D_v}$, where $D_v$ is the dimension of the visual token. After obtaining these visual tokens $\mathbf{v}_v$, our goal is to reduce the number of visual tokens and project them into language token space by a projector $\mathcal{F}(\cdot)$. As a result, the projected visual tokens will be ${\mathbf{v}_l}=\mathcal{F}(\mathbf{v})\in \mathbb{R}^{k\times D_l}$, where $k$ indicates the reduced number of visual tokens and $D_l$ is the dimension of the language tokens. By reducing the redundant visual tokens, the LLM decoder will be more efficient when performing the auto-regression since the visual tokens take a large portion of the total inputted tokens. Our primary goal is to design an efficient token reduction projector $\mathcal{F}(\cdot)$ for VLLMs, that optimizes performance during both training and inference phases. 

\begin{figure}[t]
    \centering
    \includegraphics[width=1\linewidth]{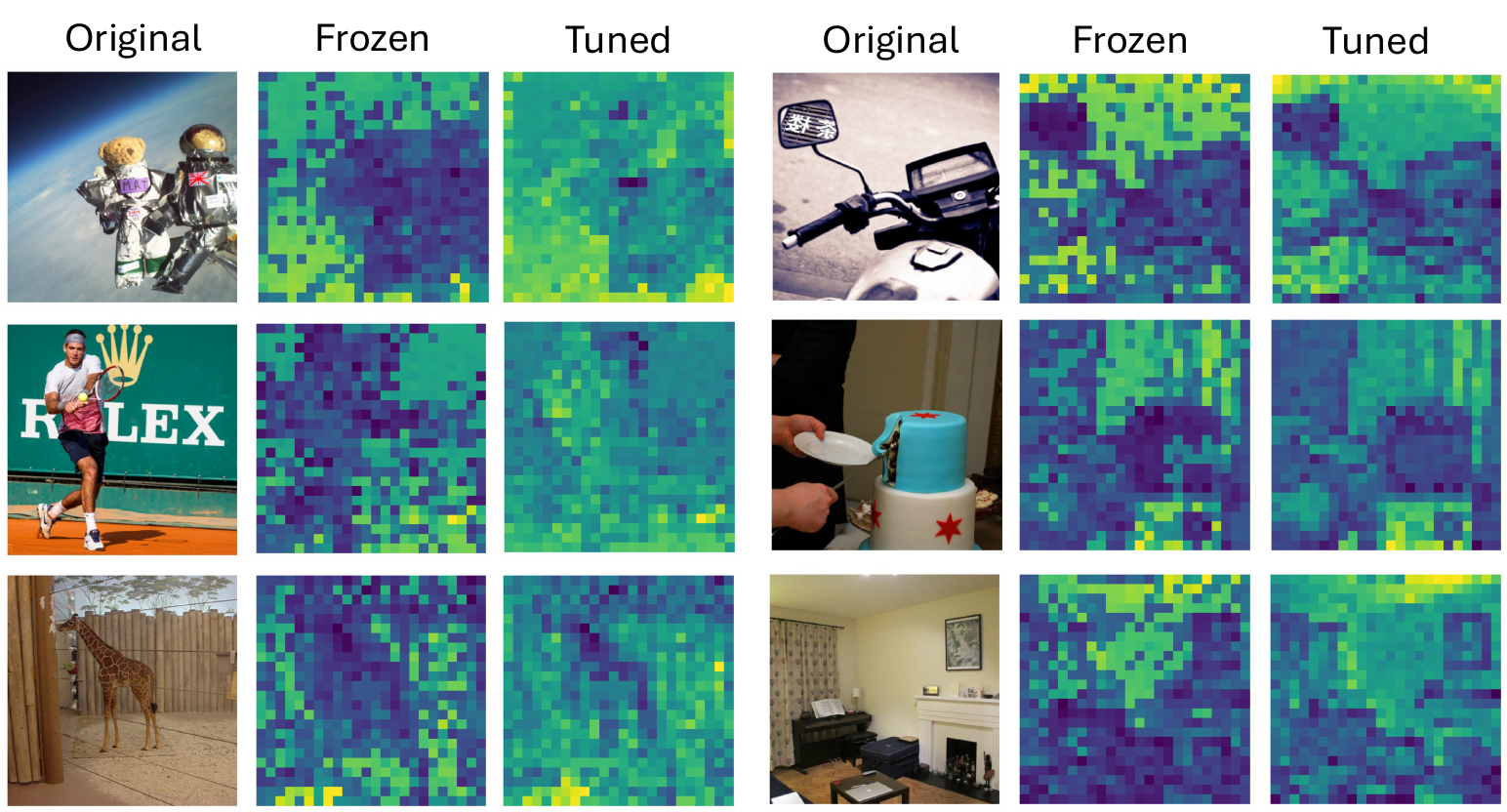}
    \caption{The visual feature map (mean pooling) comparison on LLaVA-1.5, obtained from the pretrained CLIP vision encoder by tuning the last few layers (right) and freezing all layers (middle). The tuned CLIP can learn \textit{smoother} features than the frozen one, indicating that the tokens are similar in the sliding window.}
    \label{fig:tuned_frozen}\vspace{-10pt}
\end{figure}

\subsection{Window token concatenation projector}
Neighbor patch concatenation projector in MiniGPT4-V2 \cite{chen2024minigpt} shows its effectiveness and efficiency in several VLLMs \cite{dong2024internlm,fan2024mousi} by grouping patch tokens using row-major raster scan. However, it fails to consider the spatial locality correlation of the patch tokens and also suffers the information loss issue caused by the fixed grouping strategy. Although cross-attention-based projectors like perceiver \cite{alayrac2022flamingo} or Q-former \cite{li2023blip} are adaptive in producing patch tokens with any sequence length, they will lose more information than concatenation-based ones, especially for fine-grained understanding tasks like grounding. Similarly, for selection-based projectors like token filter \cite{liu2024textmonkey} or merging-based projectors like token mixer \cite{tolstikhin2021mlp} or C-Abstractor \cite{cha2023honeybee}, they still encounter the information loss issues. Our rationale for designing the projector is to solve the drawbacks of these projectors.

As shown in \cref{fig:framework}, to minimize information loss during concatenation, we apply self-attention to all visual tokens. Specifically, the last $K_v$ blocks of a pretrained vision encoder are used as the self-attention layer. For one thing, the prior knowledge in the pretrained vision encoder offers a better initialization  for adjusting the visual tokens. For another, the self-attention layer helps capture similar visual representations within the concatenation window while distinguishing those across different windows. As shown in \cref{fig:tuned_frozen}, the feature map extracted by a tuned self-attention layer exhibits greater local similarity and smoothness compared to the one based on a frozen layer. The smoothed feature map ensures that the similar tokens are grouped within the window.
Given the visual tokens $\mathbf{v_0}$ obtained by the previous layers of the frozen vision encoder, the adjusted visual tokens ${\mathbf{v}}$ will be calculated by: ${\mathbf{v}} = \rm{SelfAttention}(\mathbf{v}_0)$, where $\rm{SelfAttention}$ is initialized by the last $K_v$ layers of the vision encoder. 

To address the inflexibility and locality-correlation problems of concatenation-based projectors, we propose a new technique based on a 2D window to group patch tokens. For adjusted visual patch tokens $\mathbf{v}\in \mathbb{R}^{n\times D_v}$, we reshape them into a 2-D feature map $\mathbf{v}'\in \mathbb{R}^{h\times w\times D_v}$, where $n=h\times w$, $h$ and $w$ indicate the height and width of the visual feature map, respectively. Typically, we set $h=w$ since the input images are usually resized as the square shape. Then we use a dynamic sliding window to turn it into output map tokens with any size. Assume the output size of the visual feature map is $h'\times w'$, where $1\le h'\le h, 1\le w' \le w$. The window size $(W_h, W_w)$ and step size $(S_h, S_w)$ are:
\begin{equation}\small
\begin{split}
    & S_h = \lfloor h/h' \rfloor, \;\;\; S_w = \lfloor w/w' \rfloor, \\
    & W_h = h - (h'-1)\cdot S_h, \; W_w = w - (w'-1)\cdot S_w,
\end{split}
\end{equation}
where $\lfloor\cdot\rfloor$ denotes the floor function. The tokens in the window will be concatenated together, and the output concatenated window tokens will be $\mathbf{v}''\in\mathbb{R}^{h'\times w'\times D_t}$, where $D_t=W_h\cdot W_w\cdot D_v$. Then we flatten the 2-D window token map into a 1-D map with the size of $k \times D_t$, $k=h'\cdot w'$. Then we use an MLP to project these 1-D visual tokens into the language space:
\begin{equation}
    \mathbf{v}_l=\texttt{MLP}(\mathbf{v}''),
\end{equation}
where $\mathbf{v}_l \in \mathbb{R}^{k\times D_l}$ are the context visual tokens for decoding. 

By using adaptive sliding window concatenation, we can preserve more visual token information which is beneficial for visual token up-sampling. Also, the sliding window will keep the spatial locality coherent based on the prior that spatially neighbored patches have similar representations. 

\begin{figure}
    \centering
    \begin{subfigure}[b]{0.23\textwidth}
        \centering
        \includegraphics[width=0.75\textwidth]{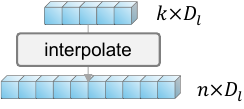}
        \caption{Token interpolation.}
        \label{fig:interp}
    \end{subfigure}
    \begin{subfigure}[b]{0.23\textwidth}
        \centering
        \includegraphics[width=0.9\textwidth]{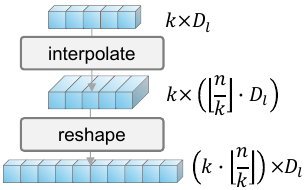}
        \caption{Channel interpolation.}
        \label{fig:interpresh}
    \end{subfigure}
    \caption{Visual token decomposition strategies by upsampling from  (a) number and (b) channel dimension. }
    \label{fig:upsample}
    \vspace{-0.5em}
\end{figure}

\subsection{Visual token decomposition}\label{sec:upsampling}
As previously discussed, we find that the number of visual tokens significantly impacts fine-grained visual understanding tasks more than coarse-grained ones. Based on this finding, we argue that the number of visual tokens impacts the perception field of the LLM. For instance, compared to a small image, a large one will be more helpful for VLLMs to recognize fine-grained objects due to the larger perception field. This phenomenon has also been demonstrated by other literature \cite{liu2023improved,li2024mini,li2023otterhd, li2023monkey}. However, increasing visual numbers means higher computation resources, and there's a trade-off between visual numbers (performance) and computation cost. To enlarge the visual tokens without inducing too much computation cost, we propose to decompose the visual tokens in the later layers of the LLM. Benefiting from the window token concatenation strategy, more information will be preserved when up-sampling the visual tokens.

Specifically, as illustrated in \cref{fig:upsample}, we explore two different up-sampling strategies for visual tokens: (a) interpolating the visual token sequence (see \cref{fig:interp}), and (b) interpolating the token channel and then reshaping the sequence (\cref{fig:interpresh}). Let $\mathbf{v}_l^{L_l-K_l} \in \mathbb{R}^{k\times D_l}$ denote the visual tokens in the ($L_l-K_l$)-th layer of the LLM decoder, where $L_l$ is the total number of the LLM layers, $K_l$ is the number last LLM layers that process the up-sampled visual tokens. We define the function $\texttt{interp}(\textbf{x},n)$ as the interpolation function over the 1st dimension of the input $\textbf{x}$ to the target dimension $n$. For the strategy (a), the up-sampled visual tokens $\hat{\mathbf{v}}_l^{L_l-K_l} \in \mathbb{R}^{n\times D_l}$ are given by
\begin{equation}
    \hat{\mathbf{v}}_l^{L_l-K_l} = {\texttt{interp}}({\mathbf{v}}_l^{L_l-K_l}, n).
\end{equation}
The benefits of this method lie in that, it is simple and efficient in implementation, and it is intuitively analogous to the local linearity of visual patch features in an image. In practice, we also find it achieves good performance, especially for fine-grained visual understanding tasks.

For the second case, as an alternative, our rationale is to restore the original visual token by expanding the compressed one from the channel dimension. Therefore, we choose to first interpolate the visual tokens from the channel, then expand the visual tokens by reshaping back to the token dimension:
\begin{equation}\small
\begin{split}    
    \tilde{\mathbf{v}}_l^{L_l-K_l} &= {\texttt{interp}}({({\mathbf{v}}_l^{L_l-K_l})}^\top, \lfloor \frac{n}{k} \rfloor \cdot D_l), \\
    \hat{\mathbf{v}}_l^{L_l-K_l} &= {\texttt{reshape}}((\tilde{\mathbf{v}}_l^{L_l-K_l})^\top, [k\cdot\lfloor \frac{n}{k} \rfloor, D_l]),
\end{split}
\end{equation}
where $\tilde{\mathbf{v}}_l^{L_l-K_l}\in \mathbb{R}^{(\lfloor \frac{n}{k} \rfloor \cdot D_l) \times k}$ are the visual tokens interpolated from channel dimension, and the operator $\lfloor\cdot\rfloor$ rounds the value to its lower-bound integer. Channel interpolation aims to span each visual token instead of inserting tokens between visual tokens, which can be roughly regarded as the inverse operation of the window token concatenation.

\section{Experiment}
We perform experiments on general VQA tasks \cref{sec:generalvqa} based on LLaVA-1.5 \cite{liu2023improved} and grounding tasks \cref{sec:grounding} based on Shikra \cite{chen2023shikra}. Then we provide the ablation study in \cref{sec:ablation} to validate the effectiveness of each module and analyze the sensitivity of hyper-parameters in our WiCo$+$. We reproduce and compare with other token reduction projector baselines, including token filter \cite{liu2024textmonkey}, perceiver \cite{alayrac2022flamingo}, token mixer \cite{tolstikhin2021mlp}, neighbor patch concatenation (concat.) \cite{chen2024minigpt} and C-Abstractor \cite{cha2023honeybee}. We compress the visual tokens to 1/4 of the original number, and all the models are trained with 8$\times$A6000Ada GPUs under the same setting. We set $K_v=1$ for the self-attention layer. We adopt the token interpolation up-sampling strategy and set $K_l=2$ for WiCo$+$ on two tasks.

\begin{table*}[t]
\centering
\caption{Comparison with different token reduction methods on six benchmarks. We reproduce all the token reduction results according to their open-sourced codes based on LLaVA-1.5 (7B).} 
\scalebox{0.9}{
\begin{tabular}{llll|cccccc}
\toprule
Method & LLM & \#Token & Res. & VQA\textsuperscript{v2} & SQA\textsuperscript{I} & VQA\textsuperscript{T} & POPE & MME & MMB \\
\midrule
BLIP-2 \cite{li2023blip}& Vicuna-13B &  256 &224  & 41.0 & 61 & 42.5 & 85.3 & 1293.8 & - \\
InstructBLIP \cite{dai2024instructblip} & Vicuna-7B  &  256 &224  & - & 60.5 & 50.1 & - & - & 36 \\
InstructBLIP \cite{dai2024instructblip} & Vicuna-13B &  256 &224 &  - & 63.1 & 50.7 & 78.9 & 1212.8 & - \\
Shikra \cite{chen2023shikra} & Vicuna-13B &  256 &224 & 77.4 & - & - & - & - & 58.8 \\
IDEFICS-9B & LLaMA-7B &  256 &224 & 50.9 & - & 25.9 & - & - & 48.2 \\
IDEFICS-80B & LLaMA-65B &  256 &224 & 60.0 & - & 30.9 & - & - & 54.5 \\
Qwen-VL \cite{Qwen-VL} & Qwen-7B &  1024 &448  & 78.8 & 67.1 & 63.8 & - & - & 38.2 \\
Qwen-VL-Chat \cite{Qwen-VL} & Qwen-7B &  1024 &448  & 78.2 & 68.2 & 61.5 & - & 1487.5 & 60.6 \\
\midrule
 LLaVA-1.5 (upper bound) \cite{liu2023improved} & Vicuna-7B &  576 &336 & 78.9 & 69.3 & 58.0 & 85.9 & 1501.7 & 65.7 \\
\midrule
LLaVA-1.5 + Token filter \cite{liu2024textmonkey} & Vicuna-7B &  144 &336 & 70.1 & 66.6 & 47.8 & 83.9 & 1267.9 & 58.2 \\
LLaVA-1.5 + Perceiver \cite{alayrac2022flamingo} & Vicuna-7B &  144 &336 & 72.3 & 69.7 & 51.5 & 82.6 & 1364.1 & 62.5 \\
LLaVA-1.5 + Token mixer \cite{tolstikhin2021mlp} & Vicuna-7B &  144 &336 & 73.5 & 69.5 & 50.8 & 83.3 & 1375.0 & 63.7 \\
LLaVA-1.5 + Concat. \cite{chen2024minigpt} & Vicuna-7B &  144 &336 & 76.3 & 68.7 & 54.5 & 84.7 & 1374.8 & 64.6 \\
LLaVA-1.5 + C-Abstractor \cite{cha2023honeybee} & Vicuna-7B &  144 &336 & 75.4 & 68.5 & 53.0 & 84.4 & 1430.6 & 63.5 \\
\rowcolor{purple!7} LLaVA-1.5 + WiCo  & Vicuna-7B &144 & 336 & {\textbf{76.5}} & 70.3 & 55.7 & \textbf{85.6} & 1463.4 & 64.3 \\
\rowcolor{purple!15} LLaVA-1.5 + WiCo+ & Vicuna-7B &144 & 336 & 76.3 & {\textbf{70.6}} & \textbf{56.0} & {85.2} & \textbf{1477.2} & \textbf{64.7} \\
\bottomrule
\end{tabular}}
\label{tab:general}
\end{table*}

\subsection{Results on general VQA tasks} \label{sec:generalvqa}
\textbf{Experiment settings}. Based on current widely-used VLLM LLaVA-1.5 (7B) \cite{liu2023improved}, we conduct all the experiments by replacing the original MLP connector to different token reduction projectors and upsampling the visual tokens in later decoder layers of the LLM (Vicuna 7B \cite{vicuna2023}). For a fair comparison, we follow the same training strategy as LLaVA-1.5 by pretraining the projector on 558K image-text pairs and finetuning both the projector and the LLM on 665K mixed instruction-following data. We evaluate all the models across six benchmarks, including VQAv2 (VQA\textsuperscript{v2}) \cite{goyal2017making}, ScienceQA (SQA\textsuperscript{I}) \cite{lu2022learn}, TextVQA (VQA\textsuperscript{T}) \cite{singh2019towards}, POPE \cite{li2023evaluating}, MME \cite{fu2023mme} and MMBench (MMB) \cite{liu2023mmbench}.

\textbf{Results analysis}. As shown in \cref{tab:general}, our WiCo ($+$) outperforms the other token reduction projectors on six benchmarks. From the results, we observe that the selection-based method ``token filter'' performs worse than all other methods, as it discards a significant amount of image information. Additionally, global merging methods like ``perceiver'' and ``token mixer'' tend to underperform on fine-grained tasks, likely due to the loss of patch position information. The concatenation-based method ``concat.'' and locality-merging method ``C-Abstractor'' perform well because they preserve more spatial and positional information. Compared to these baselines, our WiCo integrates the advantages of concatenation-based methods through window concatenation, and overcomes the downsides of these methods by smoothing the local token features. Furthermore, WiCo$+$ further increases the perception field of WiCo based on the visual token up-sampling strategy.  Compared to the original LLaVA-1.5 which exploits all the visual tokens, our WiCo$+$ can achieve almost the same performance on POPE, MME, and MMB, and even better on SQA using merely 1/4 visual tokens. 

Based on the analysis provided above, the following insights can be derived:
\begin{itemize}
    \item The performance on fine-grained understanding tasks varies significantly across different visual token reduction projectors. We attribute this variation to the loss of spatial and positional information when reducing the visual tokens;
    \item Visual token reduction will not have too much influence on course-grained visual understanding tasks, say common-sense-based VQA (SQA\textsuperscript{I}), hallucination evaluation (POPE), easy perception, cognition and reasoning (MME and MMB);
    \item Visual token reduction results in a greater performance decline for fine-grained tasks like detailed VQA (VQA\textsuperscript{v2}), and character recognition (VQA\textsuperscript{T}) compared to the coarse-grained ones.
\end{itemize}

\textbf{Time complexity}. We provide the training time comparison based on LLaVA-1.5 in \cref{tab:time_complexity}.  Since 3/4 of visual tokens have been dropped, comprising a large portion of the entire tokens, the training time improves by around 3
3h and 4h for the pretraining and finetuning stages, respectively. Although WiCo$+$ upsamples the visual tokens to 576 in the later layers of the LLM, compared to WiCo, the increase in training time is minimal. This is because only two layers receive 576 tokens, a small number relative to the total of 32 layers in Vicuna-7b.

\begin{table}[t]
    \centering
    \caption{Time complexity comparison based on LLaVA-1.5 (7B) implemented by 8*A6000Ada.}
    \label{tab:time_complexity}
    \scalebox{0.9}{
    \begin{tabular}{lccc}
    \toprule
     Methods & \#Tok. & Pretrain & Finetuning \\
     \midrule
      LLaVA-1.5 & 576 & 5h2m & 15h45m\\
      LLaVA-1.5 + WiCo & 144 & 1h53m & 11h32m\\
      LLaVA-1.5 + WiCo$+$ & 144 & 1h58m & 11h40m\\
      \bottomrule
    \end{tabular}
    }
\end{table}

\subsection{Results on grounding tasks}\label{sec:grounding}
\textbf{Experiment settings. } To evaluate the performance of our WiCo on fine-grained visual understanding tasks like grounding, we conduct experiments based on Shikra \cite{chen2023shikra} on Referring Expression Comprehension (REC) benchmarks \cite{kazemzadeh2014referitgame,mao2016generation}, \ie, RefCOCO, RefCOCO+/g. Following the training strategy as Shikra, we also perform the two-stage training, \ie, pretraining on large-scale reorganized data and finetuning on mixed instruction data. Same as Shikra, we tune both the connector and the LLM decoder during two stages. We trained only 24,000 steps instead of 100,000 for the first stage, considering training efficiency. We also follow the same hyper-parameter setting as Shikra for all the model training. Similar to \cref{sec:generalvqa}, we set $K_v=1$ for  self-attention layer.

\textbf{Results analysis}. From \cref{tab:grounding}, we can see that global-merging-based methods like ``token mixer'' and ``perceiver'' achieve the worst performance among all the token reduction projectors. We assume this may be caused by the destruction of the visual patch positional information, which is significant for grounding tasks. Different from general VQA tasks, the selection-based method ``token filter'' performs better than global-merging-based methods since it does not modify too much positional information. However, it still suffers from severe information loss, thus it performs worse than ``concat.'' and ``C-Abstractor''. Compared to ``concat.'', our WiCo can preserve the locality information through a sliding window,  which will benefit grounding tasks. Unlike C-Abstractor, which merges local tokens, our WiCo preserves more information by utilizing a concatenation strategy instead of merging. Furthermore, our self-attention design enhances the similarity of local features while preserving more detailed local information. Compared to the original Shikra, which utilizes 256 visual tokens, all methods show a substantial performance gap. We believe this is because the perception field is crucial for the grounding task, and reducing the number of visual tokens limits the model’s perception field. This also highlights the difficulty of reducing visual tokens for fine-grained perception tasks.
From the analysis of the grounding task, we can also draw the following insights:
\begin{itemize}
    \item Global-merging-based methods may destroy the patch positional information, leading to a huge performance decrease on grounding tasks;
    \item Reducing token numbers will have a higher impact on grounding tasks than on the general VQA tasks, which is related to the reduction of the perception field.
\end{itemize}

\begin{table*}[t]
\centering
\caption{Comparison with different token reduction methods on grounding tasks. We reproduce all the token reduction results according to their open-sourced codes based on Shikra.} 
\scalebox{0.9}{
\begin{tabular}{lll|ccc|ccc|cc}
\toprule
\multirow{2}{*}{Method} & \multirow{2}{*}{\#Tok.} & \multirow{2}{*}{Res.} & \multicolumn{3}{c}{RefCOCO} & \multicolumn{3}{c}{RefCOCO+} & \multicolumn{2}{c}{RefCOCOg} \\
 &  &  & Val & Test-A & Test-B & Val & Test-A & Test-B & Val & Test \\
\midrule
Shikra (upper bound) & 256 & 224 & 83.31 & 88.12 & 76.80 & 75.79 & 83.86 & 66.05 & 77.40 & 77.81 \\
\midrule
Shikra + Token Mixer \cite{tolstikhin2021mlp} & 64 & 224 & 21.99 & 21.35 & 21.99 & 15.53 & 15.61 & 14.95 & 15.95 & 16.01 \\
Shikra + Perceiver \cite{alayrac2022flamingo} & 64 & 224 & 29.20 & 32.63 & 26.16 & 18.49 & 26.68 & 15.79 & 20.75 & 21.06 \\
Shikra + Token filter \cite{liu2024textmonkey} & 64 & 224 & 59.70 & 51.70 & 44.95 & 44.14 & 51.33 & 35.59 & 44.14 & 44.19 \\
Shikra + Concat. \cite{chen2024minigpt} & 64 & 224 & 74.76 & 79.60 & {69.46} & 64.45 & 72.30 & 56.37 & 64.45 & 68.44 \\
Shikra + C-Abstractor \cite{cha2023honeybee} & 64 & 224 & 76.18 & 82.38 & 68.22 & 66.28 & 73.94 & 55.90 & 69.16 & 68.49 \\
\rowcolor{purple!7} Shikra + WiCo  & 64 & 224 & \textbf{79.20} & \textbf{85.20} & \textbf{71.05} & \textbf{69.26} & \textbf{77.52} & \textbf{57.64} & \textbf{71.10} & \textbf{71.03} \\
\bottomrule
\end{tabular}}
\label{tab:grounding}
\end{table*}

\subsection{Ablation study}\label{sec:ablation}
In this section, we first investigate the impact of each component in our method, followed by an exploration of the sensitivity of the hyper-parameters.
\begin{table*}[t]
    \centering
    \caption{Ablation study of different modules of WiCo ($+$) for LLaVA-1.5 on six benchmarks, including token decomposition, tuning of self-attention and adaptive window.}
    \label{tab:ablation}
    \scalebox{0.8}{
    \begin{tabular}{ccc|cccccc}
        \toprule
         token decomp. & self-attention & adaptive-window & VQA\textsuperscript{v2} & SQA\textsuperscript{I} & VQA\textsuperscript{T} & POPE & MME & MMB  \\
         \midrule
         \xmark & \xmark &\xmark   & 76.3 & 68.7 & 54.6 & 84.7 & 1374.8 & \textbf{64.7} \\
         \xmark & \xmark &\cmark   & \textbf{76.5} & 68.1 & 54.8 & 84.8 & 1380.4 & {64.5} \\
         \xmark & \cmark &\cmark   & \textbf{76.5} & 70.3 & 55.7 & \textbf{85.6} & 1463.4 & 64.3\\
         \cmark & \cmark &\cmark  & 76.3 & {\textbf{70.6}} & \textbf{56.0} & {85.2} & \textbf{1477.2} & \textbf{64.7}  \\
         \bottomrule
    \end{tabular}
    }
\end{table*}
\begin{table}[t]
    \centering
    \caption{Design choice of WiCo for LLaVA-1.5 on six benchmarks, including the tuning the last a few layers or using an extra self-attention layer and the choice of convolution or self-attention.}
    \setlength{\tabcolsep}{1.1mm}
    \label{tab:design_choice}
    \scalebox{0.8}{
    \begin{tabular}{c|cccccc}
        \toprule
         Methods & VQA\textsuperscript{v2} & SQA\textsuperscript{I} & VQA\textsuperscript{T} & POPE & MME & MMB  \\
         \midrule
         WiCo (convolution) & 76.2 & {69.7} & 53.2 & 84.9 & 1409.1 & 63.2  \\
         WiCo (extra self-att.) & {\textbf{76.7}} & 68.4 & {54.7} & {85.1} & {1435.4} & {\textbf{64.5}} \\
          WiCo (tuned) & {76.5} & \textbf{70.3} & \textbf{55.7} & \textbf{85.6} & \textbf{1463.4} & 64.3
         \\
         \bottomrule
    \end{tabular}
    }
\end{table}

\begin{figure}[t]
    \centering
    \begin{minipage}[t]{1\linewidth}
        \centering
        \begin{subfigure}[t]{0.485\linewidth}
            \centering
            \includegraphics[width=\linewidth]{figures/k_curves_MME.pdf}
            \caption{MME.\label{fig:k_MME}}
        \end{subfigure}
        \hspace{-1mm}
        \begin{subfigure}[t]{0.485\linewidth}
            \centering
            \includegraphics[width=\linewidth]{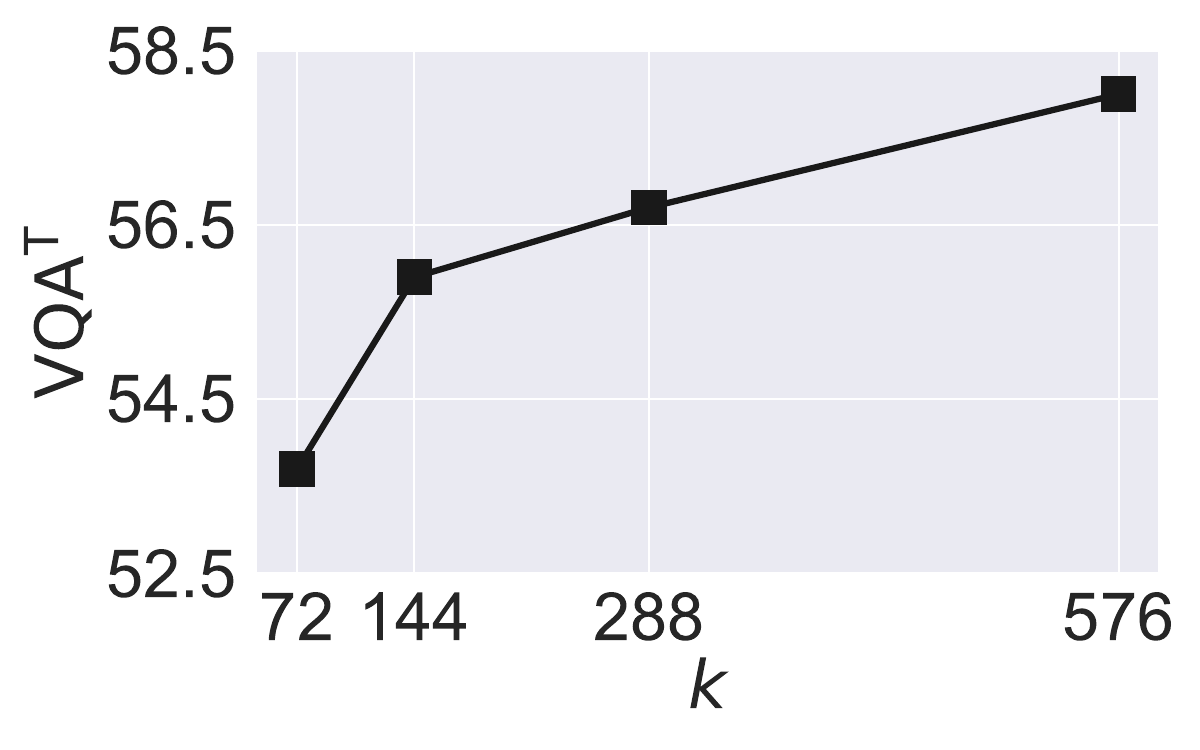}
        
        \caption{VQA\textsuperscript{T}.\label{fig:k_vqat}}
        \end{subfigure}
        \captionof{figure}{Influence of the output visual token number $k$ on MME and VQA\textsuperscript{T}.}
        \label{fig:outtokens_k}
    \end{minipage}
    \begin{minipage}[t]{1\linewidth}
        \centering
        \begin{subfigure}[t]{0.485\linewidth}
            \centering
            \includegraphics[width=\linewidth]{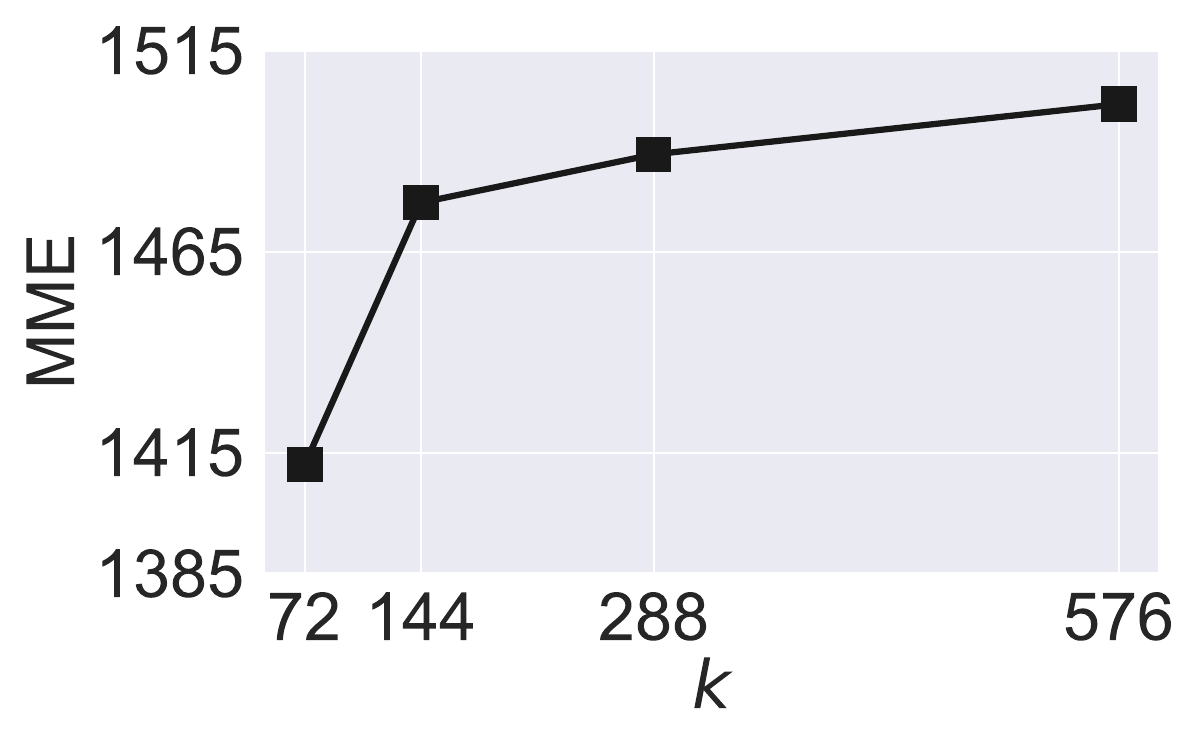}
            \caption{MME\label{fig:K_MME}}
        \end{subfigure}
        \hspace{-1mm}
        \begin{subfigure}[t]{0.485\linewidth}
            \centering
            \includegraphics[width=\linewidth]{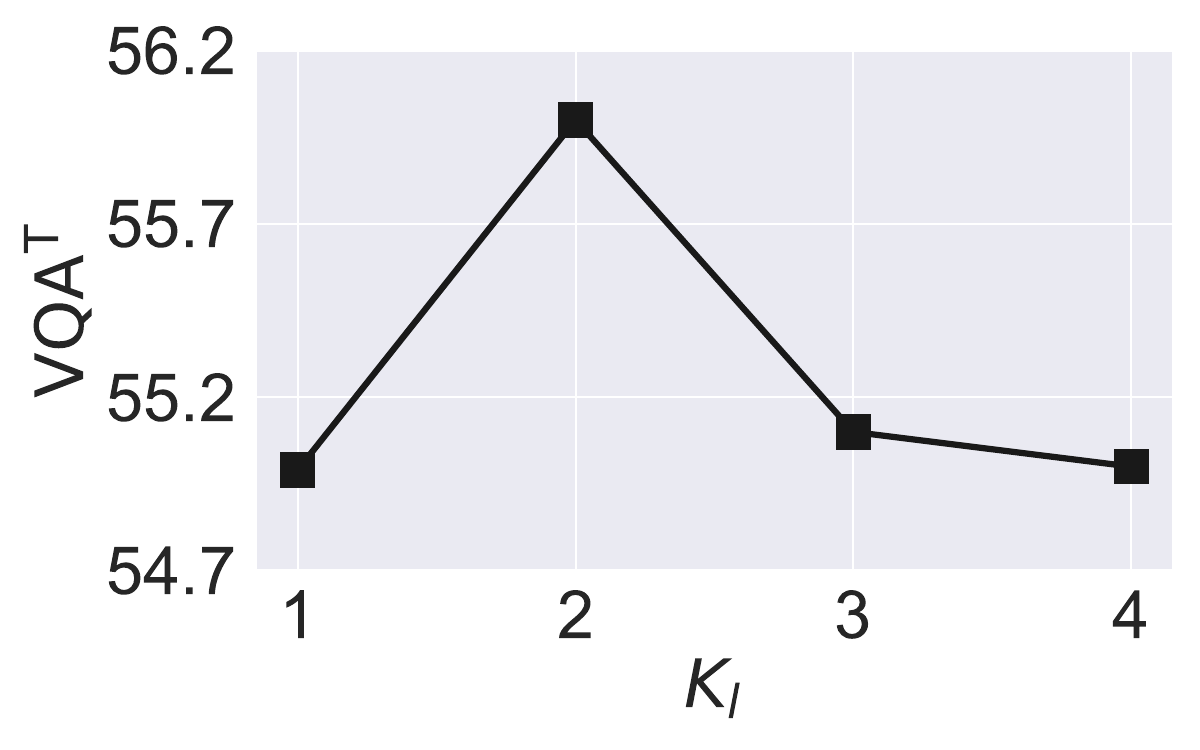}
            \caption{VQA\textsuperscript{T}\label{fig:K_vqat}}
        \end{subfigure}
        \captionof{figure}{Influence of the number of tuning layers $K_l$ on MME and VQA\textsuperscript{T}.}
        \label{fig:layer_K}
    \end{minipage}
\end{figure}

\begin{table}[t]
    \centering
    \setlength{\tabcolsep}{1.6mm}
    \caption{Comparison of two token decomposition strategies (see \Cref{sec:upsampling}), \ie, token inerpolation and channel interpolation, of WiCo$+$ for LLaVA-1.5 on six benchmarks.}
    \label{tab:interpolation}
    \scalebox{0.81}{
    \begin{tabular}{c|cccccc}
        \toprule
         Methods & VQA\textsuperscript{v2} & SQA\textsuperscript{I} & VQA\textsuperscript{T} & POPE & MME & MMB  \\
         \midrule
         WiCo$+$ (channel) & 74.9 & {68.8} & 50.1 & 82.8 & 1261.1 & \textbf{64.7} \\
         WiCo$+$ (token) & \textbf{76.3} & \textbf{70.6} & \textbf{56.0} & \textbf{85.2} & \textbf{1477.2} & \textbf{64.7} \\
         \bottomrule
    \end{tabular}
    }
\end{table}

\subsubsection{Ablation of different modules}

\textbf{The influence of the adaptive window}. As shown in \cref{tab:ablation}, using an adaptive window can improve the performance on most benchmarks compared to the unidirectional patch concatenation strategy. Specifically, for comprehensive VQA benchmarks like MME and MMB, the improvement of the adaptive window is 5.6\% and 0.8\%, respectively. This improvement is brought by the concatenation of similar spatially adjacent visual tokens, which is better than the unidirectional concatenation. Additionally, Such a window token concatenation strategy, which is based on the spatial locality proximity of visual tokens, making visual tokens within the window as similar as possible.

\textbf{The influence of the self-attention}. As shown in \cref{tab:ablation}, adding self-attention leads to improvements across most benchmarks, particularly for MME. We attribute this improvement to the aggregation of global tokens, which benefits the subsequent window patch concatenation.

Additionally, we investigate the difference of tuning last few layers and tuning an extra self-attention layer. As shown in \cref{tab:design_choice}, it can be observed that tuning last few layers of vision encoder leads to the performance improvement in most of the benchmarks especially for the TextVQA, SQA\textsuperscript{I} and MME. We assume that the visual tokens are easier to adjust based on a pretrained weight, and tuning the last few layers leverages the prior knowledge in the vision encoder. As a result, we choose to tune last few layers rather than tuning an extra self-attention layer.

Furthermore, we also consider the locality merging technique ``convolution'', by replacing the self-attention with a single 3$\times$3 convolutional layer, which is given in \cref{tab:design_choice}. We can see that the self-attention-based projector is better than the convolution-based one on most of the benchmarks. We assume this may be attributed to the benefit caused by the larger perception field. Thus, we finally  use self-attention for better aggregation.

\textbf{The influence of the token decomposition strategy}. As shown in \cref{tab:ablation}, up-sampling visual tokens in later layers of the LLM decoder can improve the performance in both fine-grained tasks and course-grained ones. Specifically, the improvement is 0.7\% on the OCR VQA benchmark and 44\% on MME, respectively. We believe this improvement stems from the larger perception field afforded by up-sampling the visual tokens. 

Furthermore, as mentioned in \cref{sec:upsampling}, we also consider interpolating channels and then expanding each token, and the comparison results are shown in \cref{tab:interpolation}. From the results, we observe that channel interpolation decreases performance, likely due to the modification of visual tokens compared to token interpolation. Consequently, we choose token interpolation as our up-sampling strategy.

\subsubsection{Hyper-parameter sensitivity}

We conduct experiments to analyze three hyper-parameters, \ie, the output token number $k$, the up-sampling layer $K_l$ and the tuning layer $K_v$. We evaluate our WiCo$+$ on a comprehensive benchmark MME and a fine-grained benchmark VQA\textsuperscript{T} by tuning these hyper-parameters. It is worth noting that for $k=576$, we utilize all the visual tokens, which result from the default configuration of LLaVA-1.5.

\textbf{The influence of the output token number $k$}. We analyze the influence of the output token number $k=\{72, 144, 288, 576\}$ in \cref{fig:outtokens_k} on MME (\cref{fig:k_MME}) and VQA\textsuperscript{T} (\cref{fig:k_vqat}), respectively. We can see from the figures that the output number of tokens has a high influence on the performance of the downstream tasks. Specifically, we observe a significant performance decline on two benchmarks when the number of visual tokens is reduced from 144 to 72. Additionally, we note that the decrease of VQA\textsuperscript{T} is greater than MME when the number of tokens changes from 576 to 144. This also indicates the insight we draw from the aforementioned experiments, \ie, visual token reduction will have a higher impact on the fine-grained tasks.


\textbf{The influence of the up-sampling layers $K_l$}.
We analyze the influence of the up-sampling layers $K_l=\{1 ,2 ,3 ,4\}$ in \cref{fig:layer_K} on MME (\cref{fig:K_MME}) and VQA\textsuperscript{T} (\cref{fig:K_vqat}), respectively. Considering the high computation cost for the LLM decoder, we only evaluate small $K_l$ in our experiments. From \cref{fig:layer_K}, we can observe that when $K_l=2$ the model reaches the best performance. It can also be seen that the variations in $K_l$ do not significantly affect the final results on both two benchmarks. Therefore, we set $K_l=2$ for all of the experiments.

\begin{table}[t]
    \centering
    \caption{Influence of the self-attention tuning layer $K_v$ of WiCo for LLaVA-1.5 on six benchmarks.}
    \label{tab:Kv}
    \scalebox{0.76}{
    \begin{tabular}{c|cccccc}
        \toprule
         Methods & VQA\textsuperscript{v2} & SQA\textsuperscript{I} & VQA\textsuperscript{T} & POPE & MME & MMB  \\
         \midrule
         WiCo ($K_v=1$) & {76.5} & \textbf{70.3} & \textbf{55.7} & \textbf{85.6} & \textbf{1463.4} & \textbf{64.3}\\
         WiCo ($K_v=2$) & \textbf{76.8} & 68.4 & 55.6 & 85.1 & 1415.3 & 63.6 \\
         \bottomrule
    \end{tabular}
    }

\end{table}

\textbf{The influence of the self-attention tuning layer $K_v$}. In our paper, we set $K_v=1$ for tuning the self-attention layer. We also try to increase the tuning layers to $K_v=2$ in \cref{tab:Kv}, but the results show that the performance will further drop on most of the benchmarks. We assume this decrease may caused by the destroy of the visual representations introducing by tuning more self-attention layers. As a result, we set $K_v=1$ in all the experiments.
\section{Conclusion}
In this paper, we investigate design choices for visual token reduction projectors in VLLMs and observe that performance on fine-grained visual understanding tasks is sensitive to the number of visual tokens. To achieve efficient visual token reduction, we introduce WiCo ($+$) and evaluate it across various benchmarks. Experiment results demonstrate the effectiveness of our approach. In the future, we believe it is a promising direction to extend our method into the video domain for better efficiency and efficacy. We hope our work can inspire more researchers to find efficient and effective token reduction projectors.


\section*{Limitations}
One limitation of our paper is the lack of experiments conducted on larger VLLMs (\eg, 13B) due to computational resource constraints. Additionally, while our adaptive window can output visual tokens with arbitrary lengths, it may result in overlapping window patches, leading to unnecessary computational costs. 
\section*{Acknowledgement}
Yifan Li, Wentao Bao, and Yu Kong are partially supported by NSF Awards 1949694 and 2040209. Any opinions, findings, and conclusions or recommendations expressed in this material are those of the authors and do not necessarily reflect the views of NSF.

{\small
\bibliographystyle{ieee_fullname}
\bibliography{egbib}
}


\end{document}